\documentclass[10pt,twocolumn,letterpaper]{article}

\usepackage{cvpr}
\usepackage{times}
\usepackage{epsfig}
\usepackage{graphicx}
\usepackage{amsmath}
\usepackage{amssymb}
\usepackage{booktabs} 
\usepackage{bm}
\usepackage{bbm}
\usepackage{subcaption}
\usepackage{caption} 
\usepackage{balance}
\usepackage{amsmath,amsfonts,amssymb}
\usepackage{amsthm}
\usepackage{booktabs}
\usepackage{algorithm}
\usepackage[noend]{algpseudocode}
\usepackage{multirow}
\usepackage{multicol}
\usepackage{threeparttable}
\usepackage[x11names,dvipsnames,table]{xcolor} 
\usepackage{colortbl}
\usepackage{algpseudocode}
\usepackage{algorithm}
\usepackage{appendix}
\usepackage{makecell}
\usepackage{diagbox}
\usepackage{url}

\newcommand\Tstrut{\rule{0pt}{2.6ex}}         
\newcommand\Bstrut{\rule[-0.9ex]{0pt}{0pt}}   

\usepackage{hyperref}
\hypersetup{pagebackref=true,breaklinks=true,letterpaper=true,colorlinks,bookmarks=false}

\cvprfinalcopy 

\def\cvprPaperID{****} 

\ifcvprfinal\fi
\begin{document}

\title{Batch-level Experience Replay with Review for Continual Learning}
\author{Zheda Mai\\
Mechanical and Industrial Engineering\\
University of Toronto\\
{\tt\small zheda.mai@mail.utoronto.ca}
\and
Hyunwoo Kim\\
AI Division\\
LG Sciencepark\\
{\tt\small eugenehw.kim@lgsp.co.kr}
\and
Jihwan Jeong, Scott Sanner\\
Mechanical and Industrial Engineering\\
University of Toronto\\
{\tt\small \{jhjeong,ssanner\}@mie.utoronto.ca}
}

\maketitle

\begin{abstract}
Continual learning is a branch of deep learning that seeks to strike a balance between learning stability and plasticity. The CVPR 2020 CLVision Continual Learning for Computer Vision challenge is dedicated to evaluating and advancing the current state-of-the-art continual learning methods using the CORe50 dataset with three different continual learning scenarios. This paper presents our approach, called Batch-level Experience Replay with Review, to this challenge. Our team achieved the 1'st place in all three scenarios out of 79 participated teams \footnote{The final ranking can be found: \url{https://sites.google.com/view/clvision2020/challenge/challenge-winners}}. The codebase of our implementation is publicly available at \url{https://github.com/RaptorMai/CVPR20_CLVision_challenge}
\end{abstract}

\section{Introduction}
\label{sec:intro}
Humans are capable of learning continuously from unlimited and non-stationary environments throughout their lives. However, without special techniques, neural networks can not learn in such manner because of \emph{catastrophic forgetting} -- the inability of a network to perform well in previously seen tasks after learning new tasks.For this reason, conventional deep learning tends to focus on offline training, where each mini-batch is sampled iid from a static dataset with multiple epochs over the training data. In recent years, \emph{Continual Learning (CL)} has received increasing attention in the machine learning community~\cite{Parisi2019}.It studies the problem of learning from a non-iid stream of data, with the goal of preserving and extending the acquired knowledge.


Current CL methods can be taxonomized into three major categories: regularization-based, parameter isolation, and memory-based methods \cite{Parisi2019}. Some regularization-based methods encode the knowledge from past tasks into a prior and utilize the prior to either regularize the update of parameters that were important to past tasks \cite{Kirkpatrick2017, SI, VCL} while others leverage knowledge distillation from the model trained on previous tasks to the model being trained on the current task \cite{Li2016, LFL}. Parameter isolation methods assign per-task parameters to bypass interference by expanding the network and masking parameters to prevent forgetting~\cite{packnet, expandable}. Memory-based methods use a memory buffer to store a subset of data from previous tasks. The samples from the buffer can be either used to constrain the parameter updates such that the loss on previous tasks cannot increase \cite{agem, gem}, or 
simply for replay to prevent forgetting~\cite{Chaudhry2019}.

Although there have been several efforts for a comparative study of continual learning, these evaluations have only been done over limited datasets and benchmarks~\cite{Parisi2019}.  The CVPR 2020 CLVision challenge on Continual Learning for Computer Vision provides an opportunity for a comprehensive comparison of state-of-the-art CL techniques~\cite{clvision2020}. 

Our approach to this challenge is based on a replay method called Experience Replay(ER), which has been shown effective in various CL problems \cite{Chaudhry2019, Aljundi2019, gss}. In the traditional ER, for every incoming mini-batch, we need to retrieve another mini-batch from the memory buffer, concatenate it with the incoming mini-batch and update the memory buffer with this incoming mini-batch. Since the data arrives in batches in this challenge, to be more efficient, we only perform the retrieval and update steps when we receive the new batch of data. Moreover, compared with ER, we add a review step before the final testing to remind the model of the knowledge it has learned during the training. 

In the following sections, we will start with a short introduction to the challenge framework, followed by a detailed description of our approach. 

\section{Challenge Framework} \label{sec:framework}
The challenge is based on the CORe50 dataset with three different scenarios and five metrics.  This section summarizes the framework of the challenge; for more details, we refer the reader to the challenge website~\cite{clvision2020}.


{\noindent \bf CORe50 dataset: }{
CORe50 is an object recognition dataset designed for different CL scenarios~\cite{Lomonaco2017}. It consists of 50 domestic objects, where each object belongs to one of the 10 categories.
The dataset has been acquired in 11 distinct sessions covering different backgrounds and lighting conditions so that different CL scenarios can be evaluated. In total, the dataset includes 164,866 RGB-D images with 128 x 128 pixels.} 


{\noindent \bf Challenge Scenarios: }{The challenge is composed of three different scenarios: {(1) New Instances}, {(2) Multi-Task New Classes}, and {(3) New Instances and Classes}. Their benchmark protocols will be described later in Section~\ref{sec:approach}.} 


{\noindent \bf Metrics: }{
To measure the performance, five metrics are evaluated as follows: 
The first two metrics evaluate classification accuracy. {(1) Final Accuracy on the Test Set} is computed once at the end of training on the hidden test set to estimate generalization performance, and 
{(2) Average Accuracy Over Time on the Validation Set} is calculated at every batch/task to measure the forgetting in the incremental setting. 
The other three metrics evaluate the computing and memory resource usage to promote the efficient methods. {(3) Total Training/Test time}, which is total running time, is measured to analyze the computational efficiency. The memory usages of RAM and disk are computed at every epoch. {(4) RAM usage} computes the total memory occupancy of the process and sub-processes, and {(5) Disk usage} calculates extra data during training time, such as replay memory and pre-trained weights.} 


\section{Approach} \label{sec:approach}
\subsection{New Instances (NI)} \label{sec:ni}
In the NI scenario, there are a total of 8 training batches presented sequentially, and each batch consists of the same 50 classes. Images within a batch are captured in similar background and lighting. The differences of background and lighting between two batches vary. Some batches are very contrasting, while some batches are quite similar. We also remark that in the NI scenario, no batch label is given during both training and testing time. 

Since the data batches are encountered over time, a naive way to train the dataset is to train each batch in an offline manner, with multiple epochs over the data and shuffling of the data to ensure they are i.i.d.\ within this batch. However, when the neural network trained on the previous batch encounters a new batch, due to the difference between the data distributions of two batches, it often experiences catastrophic forgetting \cite{Parisi2019}.

To mitigate this problem, a common approach in CL is the memory-based method described in \ref{sec:intro}, which uses episodic memory to store a subset of the data from past batches to tackle forgetting. The data from the episodic memory can be used to either constrain the optimization of parameters such that the loss on past batches can never increase~\cite{gem, agem} or conduct experience replay~\cite{Chaudhry2019, Aljundi2019}. In this competition, we chose the experience replay approach as it is more efficient and computationally cheaper compared to the constraint optimization approach.

{\noindent \bf Batch-level Experience Replay with Review: }{
Compared with the simplest baseline model that fine-tunes the parameters based on the new task without any techniques to prevent forgetting, ER stores a subset of the samples from the past batches in a memory buffer $\mathcal{M}$ of limited size $mem\_sz$. During the training of the current batch, it concatenates the incoming mini-batch with another mini-batch of samples retrieved from the memory buffer. Then, it simply takes an SGD step with the combined batch, followed by an update of the memory~\cite{Chaudhry2019}. 

Since we need to perform the retrieval and update steps for every minibatch, this approach will not be efficient when we have thousands of mini-batches. Also, as data arrives in batches in the challenge, we concatenate the memory examples at the batch level instead of concatenating the memory examples at the minibatch level. Additionally, we add a review step before the final testing to remind the model with the knowledge it has learned. 

The overall training procedure is presented in Algorithm~\ref{alg:algo1}. For every batch of data except the first batch, we do a batch level experience replay. Concretely, for every epoch, we draw another batch of data $D_\mathcal{M}$ randomly from the episodic memory with size  $replay\_sz$, concatenate it with the current batch and conduct the gradient descent parameters update. We note that $D_\mathcal{M}$ is different for every epoch.  When we finish the last epoch of the current batch, we will randomly select $\frac{mem\_sz}{n}$ examples from the current batch where $n$ is the total number of batches in the whole scenario. After training all the data batches, we will do a final review step where we draw a batch of size $D_R$ from memory and conduct the gradient update again. We note that, to prevent overfitting, the learning rate in this step is usually lower than the learning rate used when processing new batches.}

\begin{algorithm*}[t!] 
\caption{Batch-level Experience Replay}
\label{alg:algo1}
\begin{algorithmic}[1]
\Procedure{ER}{$\mathcal{D}$, mem\_sz, replay\_sz, review\_sz, batch\_sz, lr\_replay, lr\_review}
    \State $\mathcal{M}\leftarrow \{\} * mem\_sz$  \Comment{Allocate memory of size mem\_sz}
    \For{$t\in\{1,\dots,T\}$}
        \For{epochs}
            \If {$t > 1$}
                \State $D_\mathcal{M}\stackrel{\text{replay\_sz}}{\sim}\mathcal{M}$ \Comment{Sample a batch of data with size replay\_sz from $\mathcal{M}$}
                \State $D_{\text{train}} = D_{\mathcal{M}}\cup D_t$ \Comment{Concatenate the current data batch and the memory batch}
            \Else
                \State $D_{\text{train}} = D_t$
            \EndIf
            \State $\theta \leftarrow$ SGD($D_{\text{train}}, \theta, \text{lr\_replay}, \text{batch\_sz}$) \Comment{One pass minibatch gradient descent over $D_\text{train}$}
        \EndFor
        \State $\mathcal{M}\leftarrow UpdateMemory(D_t, \text{mem\_sz})$ \Comment{Update memory}
    \EndFor
    \State $D_R\stackrel{\text{review\_sz}}{\sim}\mathcal{M}$ \Comment{Sample a batch of data with size review\_size from $\mathcal{M}$}
    \State $\theta\leftarrow \text{SGD}(D_R,\theta,\text{lr\_review}, \text{batch\_sz})$ \Comment{One pass minibatch gradient descent over $D_\text{R}$}
    \State \textbf{return} $\theta$
\EndProcedure
\end{algorithmic}
\end{algorithm*}

\subsection{New Instances and Classes (NIC)} \label{sec:nic}
In the NIC scenario, there are 391 training batches containing 300 images of a single class. Similar to the NI scenario, if we just fine-tune the model with new incoming batches, the model will experience catastrophic forgetting. So, we use the same algorithm to tackle the NIC scenario as NI but the parameters used in NIC are very different than the ones we used in NI. The detailed hyper-parameters of the models for each scenario will be listed in Appendix~\ref{sec:appA}.

\subsection{Multi-Task New Classes (Multi-Task-NC)} \label{sec:multi-task-nc}
In this scenario, 50 different classes are split into 9 different batches: 10 classes in the first batch and 5 classes in the other 8. The main difference between this setting and the other two is that in this case, the task label will be provided during training and test. Therefore, the task difficulty is much smaller than the other two scenarios. 

As~\cite{mer} proposed recently, we can treat the CL problem as solving Transfer-Interference Trade-off, where we want to maximize transfer and minimize interference. In this scenario, we found that interference outweighs transfer when we share the same model across all the batches. Thus we decided to assign a fresh pre-trained model for each batch to prevent interference. Moreover, since we don't need any extra steps to avoid forgetting, we will have shorter training time as well. But the drawback of this method is that it prevents positive transfer due to a lack of weight sharing. 

\begin{figure}[h!]
    \centering
    \includegraphics[width=0.8\columnwidth]{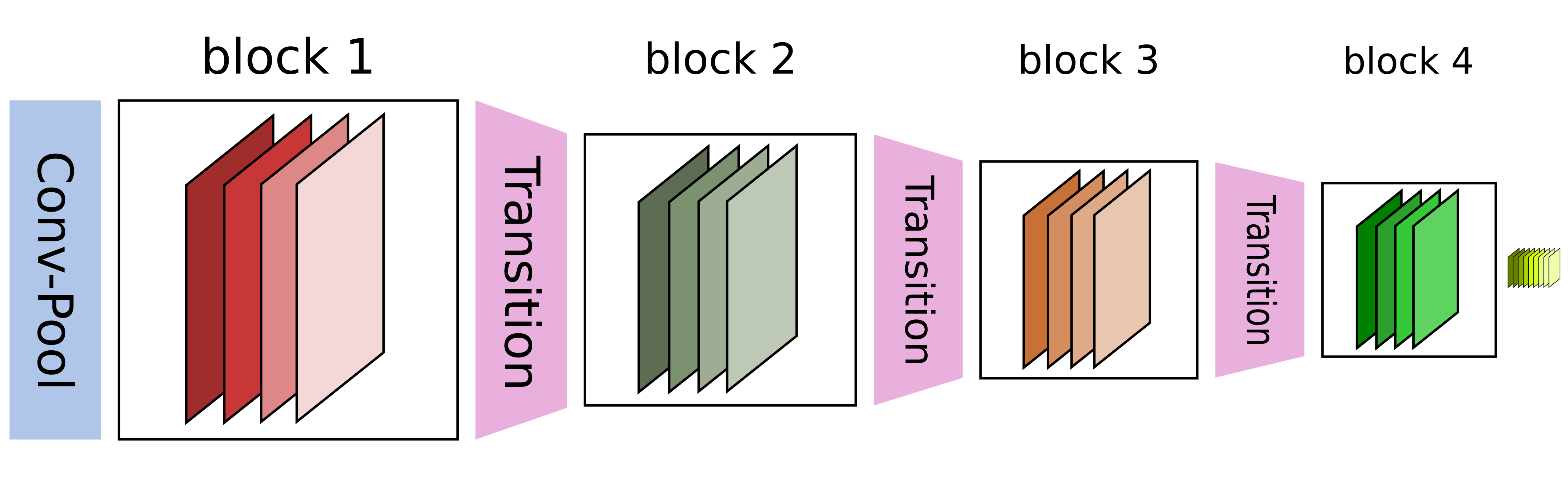}
    \caption{DenseNet-161 Architecture \protect\cite{blog}}
    \label{fig;dense}
\end{figure}

\subsection{Architecture and Training Details}\label{sec:arch}
For all three scenarios, we use the DenseNet-161 model~\cite{Huang2017} pre-trained on ImageNet. DenseNet-161 is the largest model in the DenseNet group with a size around 100MB. As shown in Figure~\ref{fig;dense}, the DenseNet-161 model consists of 4 dense blocks and we freeze all the layers before the third blocks. This ensures the pre-trained model can extract the basic features from the image and shorten the training time as well. The network is trained via cross-entropy loss and stochastic gradient descent with the mini-batch size equal to 32. The detailed hyper-parameters of the optimizer for each scenario are listed in Appendix~\ref{sec:appA}.

\begin{figure}[h]
    \centering
    \includegraphics[width=0.4\columnwidth]{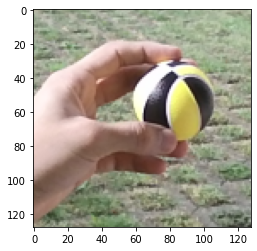}
    \includegraphics[width=0.4\columnwidth]{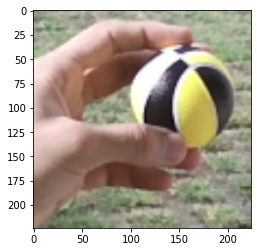}
    \caption{Example of pre-processing (e.g., center-crop)}
    \label{fig:crop}
\end{figure}

The size of an image in the challenge is (128, 128, 3). Before feeding an image to the model, we preprocess the image by center-cropping the image with size (100, 100) and resizing it to (224, 224, 3). As we mentioned in~\ref{sec:ni}, the critical differences between batches are background and lighting, and most of the target objects are in the center of the images. Therefore,  center-cropping helps mitigate the background's effect to some extent as the target object takes more space in the cropped image. As we do not train any layers before the third dense block, we resize the cropped image to the size of (224, 224, 3) to ensure no size discrepancy between the pre-trained model and the training images. Noted that this preprocessing step is applied to both training and testing images. Figure~\ref{fig:crop} shows an example of the center-cropped image.

Moreover, to get better generalization, we leverage data augmentation techniques, including pixel-level and spatial-level transformations. Specifically, we use five spatial-level transformations, including HorizontalFlip, RandomRotate90, ElasticTransform, GridDistortion, and OpticalDistortion. For pixel-level transformations, we use RandomContrast, RandomGamma, and RandomBrightness. The details of the data preprocessing steps used in the final submission are listed in Appendix~\ref{sec:appB}.

\section{Experiment} \label{sec:exp}
\subsection{Multi-Task New Classes (Multi-Task-NC)}
The Baseline method in Multi-Task-NC shares all the layers of the model before the last fully-connected layer between all the batches and each batch has its own fully connected layer. Ind\_model represents the independent model approach mentioned in~\ref{sec:multi-task-nc} that a fresh pre-trained model is assigned to each batch of data. Ind\_model\_preproc is the Ind\_mdoel plus the preprocessing steps mentioned in~\ref{sec:arch}. DenseNet161\_tune\_all means tuning the model without freezing any of the layers and DenseNet161\_freeze freezes all the layers before the third block as mentioned in~\ref{sec:arch}.

As we can see in Table~\ref{table:nc}, the baseline method experiences huge forgetting. By assigning an independent model to each batch and freezing the first half of the model, the performance improves significantly and the processing step helps the model generalize better. 
\begin{table}[ht!]
\caption{Multi-Task-NC Performance Comparison}
  \rowcolors{2}{gray!10}{white}
  \resizebox{\linewidth}{!}{%
  \begin{tabular}{cccc}
  \toprule
    Method & Architecture &  avg\_val\_acc & final\_val\_acc \\ 
  \midrule
    Baseline & DenseNet161\_tune\_all & 14.0\% & 12.8\%\\
    Ind\_model & DenseNet161\_freeze&54.7\%&98.6\%\\
    Ind\_model\_preproc&DenseNet161\_freeze&55.1\%&99.3\%\\
    \hline
  \end{tabular}
  }
 \label{table:nc}
\end{table}

\subsection{New Instances (NI)}
In the NI scenario, all batches share the same model and the baseline is fine-tuning the model based on new incoming batches without any steps to prevent forgetting. The Batch-level Experience Replay (BER) improves the final validation accuracy by around 7.7\%. The final review step increases the final validation accuracy by 1.1\% but it does not help the average validation accuracy since the final review is done by the end of the training. The data preprocessing and augmentation yield much better generalization and consequently enhance both metrics. 
\begin{table}[ht!]
\caption{NI Performance Comparison}
  \rowcolors{2}{gray!10}{white}
  \resizebox{\linewidth}{!}{%
  \begin{tabular}{cccc}
  \toprule
    Method & Architecture &  avg\_val\_acc & final\_val\_acc \\ 
  \midrule
  Baseline & DenseNet161\_tune\_all & 71.1\% & 81.1\%\\
  BER & DenseNet161\_tune\_all & 77.1\% & 88.7\%\\
  BER\_review & DenseNet161\_tune\_all & 77.0\% & 89.8\%\\
  BER\_review\_preproc & DenseNet161\_freeze &90.1\%& 96.7\%\\
    \hline
  \end{tabular}
  }
 \label{table:ni}
\end{table}

\subsection{New Instances and Classes (NIC)}
The baseline of NIC is the same as NI. However, since in NIC, every batch contains only one class, the data distribution difference between two batches is much greater than the NI scenario, which explains the extremely poor result of the baseline. We use the same algorithm, Batch-level Experience Replay with Review, to tackle this scenario as well. Similar to the result of NI, the BER\_review\_preproc method gets the highest values in both metrics. The average validation accuracy of NIC is much lower than NI. This is because we capture validation accuracy by the end of each batch and at the beginning of the training, the validation accuracy is very low as the model has not seen enough data yet. 
\begin{table}[ht!]
\caption{NIC Performance Comparison}
  \rowcolors{2}{gray!10}{white}
  \resizebox{\linewidth}{!}{%
  \begin{tabular}{cccc}
  \toprule
    Method & Architecture &  avg\_val\_acc & final\_val\_acc \\ 
     \midrule
    Baseline &DenseNet161\_tune\_all & 0.02\% & 0.02\%\\
    BER\_review & DenseNet161\_tune\_all & 55.3\% & 90.1\%\\
    BER\_review\_preproc & DenseNet161\_freeze &59.4\%& 96.0\%\\
    \hline
  \end{tabular}
  }
 \label{table:nic}
\end{table}

\section{Conclusion} \label{sec:conc}
In this paper, we described our approach for the CVPR 2020 CLVision  Continual Learning for Computer Vision challenge. Compared with the traditional ER, the proposed Batch-level Experience Replay with Review makes two modifications: (1) it retrieves samples from memory when the model receives a new batch of data and updates the memory after training the current batch to reduce the total number of memory retrieval and update steps; (2) it performs a review step before the final testing to remind the model of the knowledge it has learned during the whole training.

In NI and NIC scenarios, the proposed method improved the final validation accuracy and the average validation accuracy by large margins in comparison to the baseline method.  In the Multi-Task-NC scenario, fine-tuning a fresh pre-trained model for each batch turns out to be a simple but effective approach. Overall, our approaches achieved highly competitive performance and won the 1'st place in all three scenarios out of 79 teams. 
\section*{Acknowledgement}
We thank our colleague Ga Wu who provided insight and expertise that greatly helped us during the challenge.

\clearpage
{\small
\bibliographystyle{ieee_fullname}
\bibliography{egbib}
}
\newpage

\onecolumn

\appendix
\renewcommand*\appendixpagename{\large \textbf{Appendix}}
\appendixpagename

\section{Hyper-parameters of the model training}\label{sec:appA}

\begin{table}[h!]
\centering
    \normalsize
    \begin{tabular}{c|c|c|c}
    \diagbox{\textbf{Parameter}}{\textbf{Scenario}} & NC & NI & NIC\\
    \hline \Tstrut
    Method & Ind\_model\_preproc & ER\_review\_preproc & ER\_review\_preproc \\[0.5ex]
    \hline \Tstrut
    Optimizer & SGD & SGD & SGD \\[0.5ex]
    \hline\Tstrut
    Batch size & 32 & 32 &32 \\ [0.5ex]
    \hline\Tstrut
    Preload data & No & No & No \\[0.5ex]
    \hline\Tstrut
    \# Replay examples & -& 10000 & 200 \\[0.5ex]
    \hline\Tstrut
    \# Replay used & -& 10000 & 600 \\[0.5ex]
    \hline\Tstrut
    Epochs & 1 & 2 & 2 \\[0.5ex]
    \hline\Tstrut
    Review size & -& 20000 & 20000 \\[0.5ex]
    \hline\Tstrut
    Review epoch &- & 1 & 1\\[0.5ex]
    \hline\Tstrut
    Review lr decay factor& -&0.5&0.5\\[0.5ex]
    \end{tabular}
    \caption{Hyper-parameters of the model training.}
\end{table}

\section{The data preprocessing steps}\label{sec:appB}
\normalsize
\begin{table}
\centering
\begin{tabular}{  c | c | c }
  \thead{Step} & \thead{Augmentation} & \thead{Probability}\\[0.5ex]
  \hline\hline \Tstrut 
  Step 1 & CenterCrop(100, 100) & $p=1.0$ \Bstrut\\
  \hline \Tstrut
  Step 2 & One of $\begin{cases} \text{HorizontalFlip}(p=1) \\ \text{RandomRotate90}(p=1) \end{cases}$ & $p=0.5$ \Bstrut\\
  \hline
  Step 3 & One of $\begin{cases} \text{RandomContrast}(0.4) \\ \text{RandomGamma}(20, 180)\\ \text{RandomBrightness}(0.4) \end{cases}$ & $p=0.5$\\
  \hline
  Step 4 & One of $ \begin{cases} 
  \text{ElasticTransform}(\alpha=120, \sigma=6, \alpha_{\text{affine}}=3.6) \\ \text{GridDistortion}()\\ 
  \text{OpticalDistortion}(\text{distort\_limit}=2, \text{shift\_limit}=0.5) \end{cases}$ & $p=0.3$ \\
  \hline \Tstrut
  Step 5 & Resize(224, 224) & $p=1.0$\Bstrut\\
  \hline
  Step 6 & Normalize$\bigg(\begin{array}{l} mean=[0.485, 0.456, 0.406],\\std=[0.229, 0.224, 0.225]\end{array}\bigg)$ & $p=1.0$ \\
\end{tabular}
\caption{The details of the data preprocessing steps.}
\end{table}

\rotatebox{90}

\end{document}